\documentclass[sigconf]{acmart}
\acmConference[4th CSAI]{CSAI}{December 2024}{Beijing, China}

\AtBeginDocument{%
  }

\begin{document}

\title{Towards Selection and Transition Between Behavior-Based Neural Networks for Automated Driving}

\author{Iqra Aslam}
\email{iqra.aslam@tu-clausthal.de}
\affiliation{%
  \institution{Institute for Software and System Engineering, TU Clausthal}
  \postcode{38678}
  \country {Germany}
}

\author{Igor Anpilogov}
\email{igor@igisaur.com}
\affiliation{%
  \institution{Institute for Software and System Engineering, TU Clausthal}
  \postcode{38678}
  \country {Germany}
}
\author{Andreas Rausch}
\email{andreas.rausch@tu-clausthal.de}
\affiliation{%
 \institution{Institute for Software and System Engineering, TU Clausthal}
 \postcode{38678}
 \country {Germany}
 }

\renewcommand{\shortauthors}{Aslam et al.}

\begin{abstract}
Autonomous driving technology is progressing rapidly, largely due to complex End-To-End systems based on deep neural networks. While these systems are effective, their complexity can make it difficult to understand their behavior, which raises safety concerns. This paper presents a new solution: a Behavior Selector that uses multiple smaller artificial neural networks (ANNs) to manage different driving tasks, such as lane following and turning. Rather than relying on a single large network, which can be burdensome, require extensive training data and is hard to understand. The developed approach allows the system to dynamically select the appropriate neural network for each specific behaviour (i.e. turns) in real time. We focus on ensuring smooth transitions between behaviors and taking into account the vehicle’s current speed and orientation to improve stability and safety. The proposed system has been tested using the AirSim simulation environment, demonstrating its effectiveness. 
\end{abstract}

\keywords{E2E learning,  Behavior based neural networks, Behavior based control, Multi-State system, Behaviour selector, Behavior transition}

\maketitle

\section{Introduction}
The field of autonomous driving is undergoing a stage of rapid advancements. There are a vast number of approaches explored to develop the most effective system for controlling vehicles without human intervention. The primary goal is to navigate the vehicle from its current position to a specified target location autonomously. There are two main approaches to achieving this: the classical modular pipeline approach and the End-to-End (E2E) learning approach \cite{tampuu2020survey}. The modular pipeline breaks down the autonomous driving task into sub-tasks such as perception, planning, and control \cite{aslammethod}. In contrast, the E2E approach replaces the entire pipeline with a single, large artificial neural network (ANN) that learns to map raw sensor data directly to vehicle control signals. While the E2E approach has shown promise, especially with the development of efficient ANNs, one of the biggest breakthroughs is the creation of efficient ANN. It required vast amounts of training data and time to train a good network that could handle all driving maneuvers (e.g., lane following, turning, stopping) \cite{chen2024end}. Moreover, these networks can be difficult to interpret and may behave unpredictably in unforeseen situations.

To mitigate these issues, one solution is to use multiple smaller ANNs, each responsible for a specific driving behavior (e.g., lane following) rather than relying on a single network for the entire driving task. This approach requires less training data, simplifies the network, and makes training faster and easier to validate. For instance, a network dedicated solely to lane following does not need to be trained and validated on data that includes all driving behaviors, instead just use lane-following data.

However, managing multiple ANNs introduce new challenges: (1) only one network can control the vehicle at any given time, necessitating an arbiter or Behavior Selector (BS) to activate the appropriate driving behavior. (2) Ensuring smooth switching between different driving behaviors requires additional mechanisms. (a) the current selected behavior then should correct the state (e.g., speed and orientation to be needed for the next driving sequence) of the vehicle in such a way that the next behavior can operate. For example, consider the switch between the Follow Lane behavior and the Turn Left behavior. Turn Left behavior is operating at low speed, while Follow Lane is operating at high speed. If the vehicle accelerates too high quickly while following a lane, it may not be able to safely execute a sharp turn as trained (b)additionally, if the vehicle’s orientation is misaligned with the lane at the time of switching. Since the switch occurs at a semi-random position on the road, the previous driving function may not have completed its maneuver, leaving the vehicle misaligned with the lane, it may deviate from the intended path. This could be problematic, as the training data typically assumes the vehicle is parallel to the lane. This aligns with the “\emph{Garbage In, Garbage Out}" principle \cite{witten2002data}: if the vehicle’s state doesn’t match the training conditions, the behavior will be incorrect. This issue underscores the importance of aligning the vehicle’s state with the training data to prevent suboptimal performance.

This paper investigates solutions for integrating a behavior-based approach with neural network-driven systems. It aims to identify methods by which the Behavior Selector can effectively select and transition between E2E behaviors based on the route planner’s driving commands while addressing the challenges outlined.

The rest of this paper is organized as follows: Section II provides an overview of related work in E2E driving and behavior selectors, focusing on both robotics and automotive applications. Section III presents the system architecture and different conceptual approaches for the behavior selector. In Section IV, Evaluation and discussion of the results are presented. Section V concludes the paper by summarizing its contributions and suggesting potential directions for future research.

\section{Related Work}
The concept of creating autonomous driving vehicles using ANNs was first introduced by D. A. Pomerleau in 1990 \cite{pomerleau1988alvinn}. The ANNs take a large amount of training data as well as computational power. Despite having these disadvantages, the neural network approach proved to be the most effective at mimicking the driving behavior of a human driver. However, given the complexity of driving, it’s impractical to train a single network to handle all possible scenarios. This led to the development of behavior-specific E2E models, where individual distinct networks tackle specific tasks. It can be described as a multi-state system or as “\emph{state of a system}" or “\emph{system behavior}", a concept explored in 1971 \cite{ackoff1971towards} which viewed complex systems as collections of distinct operational states. 

Building on this, \cite{murchland1975fundamental} J.D. Murchland, contributed to understanding how systems switch between behaviours, particularly focusing on reliability. Murchland's work is particularly relevant when considering the design and evaluation of autonomous driving systems that rely on multiple behavioral states to manage complex driving tasks. While his focus was on the reliability aspects of these transitions, lacking the real-time operation of autonomous vehicles.

As the field evolved, the need for a control mechanism to switch between different behaviors in a multi-state system became apparent. In 1986, R. A. Brooks \cite{brooks1986robust} laid the foundation for these control systems in robotics, particularly with the use of state machine-based mechanisms for task-achieving behaviors. As such, the decomposition of a mobile robot control system based on task-achieving behaviors greatly influenced the design of behavior selectors, which manage the transition between different driving behaviors in autonomous systems.

Later in 1997 \cite{rosenblatt1997damn} J.K. Rosenblatt applied the ideas of DAMN (Distributed Architecture for Mobile Navigation) to autonomous vehicles and robotics, that focused on task-oriented behaviors, i.e. using an arbiter to switch between different navigational patterns like obstacle avoidance, goal-seeking, and path-following. The DAMN architecture utilized an arbiter to manage and prioritize these behaviors, allowing the system to make real-time decisions by blending the outputs of various behaviors through a voting mechanism. However, the behaviors were more generalized and focused on fundamental navigational tasks rather than specific driving maneuvers like turning left, turning right, or lane following, which are crucial for autonomous driving. Additionally, unlike modern behavior selectors, DAMN did not consider the vehicle's dynamic state (e.g., speed and orientation) during behavior transitions, making it less suitable for the sophisticated demands of E2E autonomous driving.

A related behavior-based approach was employed \cite{langer1994behavior} to address specific autonomous tasks relevant to off-road environments. The system utilized an arbiter to manage different behaviors. Like \cite{rosenblatt1997damn} this study has also focused on basic navigational tasks in unstructured, rugged environments. Each behavior had a specific function, and the arbiter would combine the outputs from these behaviors using a weighted sum to decide the vehicle's next action. The focus was on ensuring the vehicle could navigate complex off-road terrains by dynamically responding to immediate environmental challenges, rather than following predefined paths or performing precise maneuvers like lane following or making sharp turns, which are more typical in structured, on-road settings.

More recently, Orzechowski et al. \cite{orzechowski2020decision} described a hierarchical decision-making system for autonomous vehicles, which integrated multiple layers of behavioral patterns into a scalable framework. Their work focused on handling various driving scenarios on combining basic systems to form complex decisions to handle various driving scenarios. It did not incorporate machine learning, a core component of the behavior selector in this paper. Machine learning adds complexity to the system, particularly in managing transitions between behavioral patterns, which must consider the vehicle’s state at the transition point. This oversight can lead to challenges in ensuring smooth and safe transitions between behaviors, a critical aspect of autonomous driving systems. Additionally, their system did not incorporate a Route Planner for short-term goal-based navigation, which is essential for effectively managing driving tasks and transitions in current behavior-based systems.


Despite the advances made in these areas, most existing studies still rely heavily on rule-based methods and lack mechanisms for seamless behavioral transitions in dynamic environments. This is where the fuzzy logic-based approach by Fouladinejad \cite{fouladinejad2015towards} adds value, enabling smoother decision-making by accounting for uncertainties in real-world driving. Building on these foundations, our approach advances these ideas by integrating AI based driving behavior and manage transitions between basic driving behaviors while considering the dynamic state of the vehicle.


\section{Driving behavior selector: system Architecture and concept:}

This section initially describes the system architecture for a behavior-based E2E autonomous driving system, followed by different solutions for implementing and enhancing the Behavior Selector, which addresses the challenges stated in the introduction.


\begin{figure*}[ht]
\centering
\includegraphics[width=0.65\textwidth]{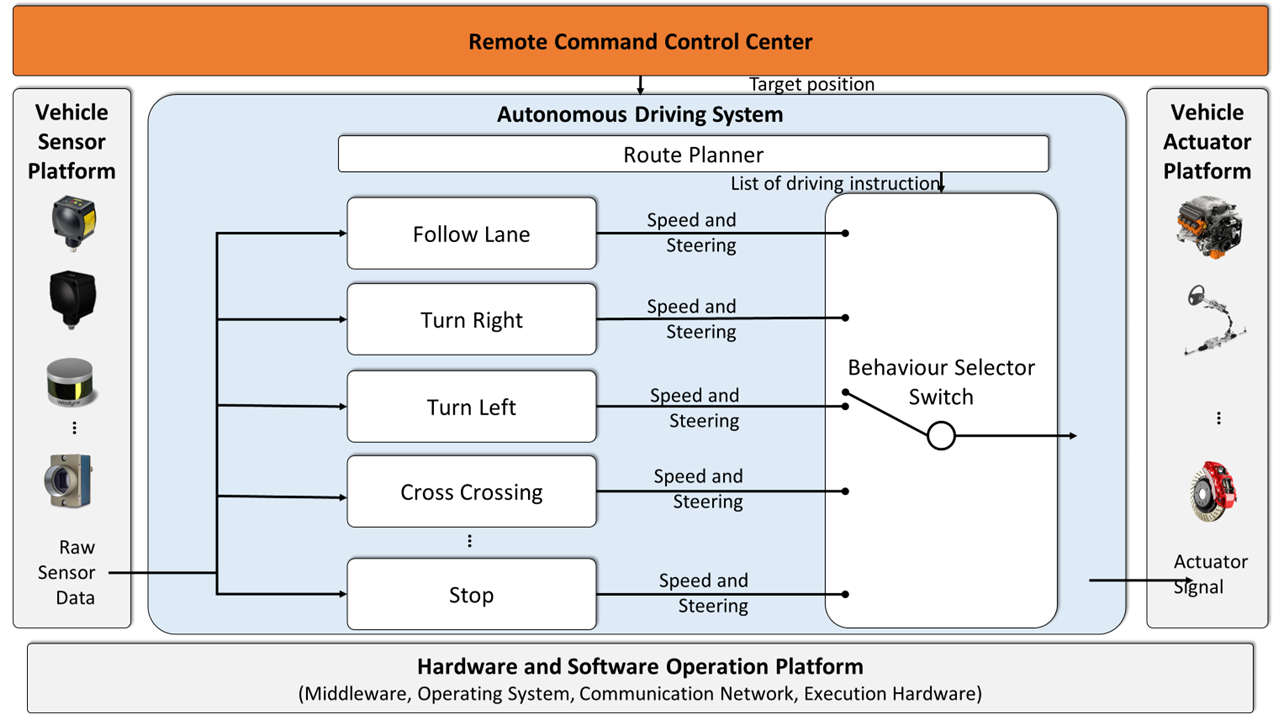}
\caption{Architecture for a behavior-based E2E autonomous driving system}\label{architecture}
\end{figure*}

Figure \ref{architecture} describes the system architecture for a behavior-based E2E autonomous driving system. The architecture consists of three main components: (1) a Vehicle sensors platform, (2) driving behavior-specific neural networks, a route planner, and a behavior selector, and (3) a Vehicle actuator platform to control the vehicle. On the left side, the sensor perceives the environment and provides the sensory input to the E2E behaviors, while the Route Planner gets the current position of the vehicle, orientation, and target position from some external system named “Remote command control center” \cite{aniculaesei2023connected} \cite{aslam2024runtime}. 
The route planner is an important sub-component based on the Open-DRIVE format\cite{opendrive:online}. It calculates the path to reach the destination and sends it in the form of a sequence of driving instructions or behaviors to the Behavior Selector. The Behavior Selector uses this information as input to choose the appropriate behavior (e.g., Follow Lane or Turn Left). Although the route planner’s detailed implementation is beyond the scope of this paper. In the middle, multiple neural network modules act as driving behaviors. Each network is trained to handle specific driving tasks, such as lane following, making turns, stopping after reaching the destination, and lastly, crossing the crossing. Cross the crossing is considered as a separate behavior as it is different than follow lane behavior and we don’t have any lane to follow as in the case of follow lane behavior (e.g. junctions).  

The outputs from all the ANNs consist of the desired speed and steering values and are sent to the Behavior Selector. The Behavior Selector is the core component and contribution of this paper. It receives the sequence of driving instructions and the distance to them from the route planner. Using these, the Behavior Selector selects the required neural network behavior and forwards the control signals of respective behavior (speed and steering) to actuator sensors to control the vehicle.

To implement the behavior selector that ensures a safe and smooth transition among driving E2E networks four different approaches have been proposed. First, the Basic Behavior Selector is based on route planner inputs without any additional transition and interpolation. The second Behavior Selector is based on the transition behavior approach that includes creating an additional driving behavior to adjust the vehicle’s state e.g., if the vehicle is accelerating during Follow Lane behavior, the Transition behavior will reduce the speed before a turn, ensuring the vehicle's state matches the requirements of the next driving behavior network, ensuring smoother changes between different driving behaviors. Third is an interpolation-based approach that interpolates between the driving behavior, to blend the outputs of the current and next networks to create a more gradual change and improve stability. Fourth, the Behavior Selector is based on Transition and Interpolation, a hybrid approach that combines both transition and interpolation for the most seamless behavior switching. Before delving into a detailed discussion of each approach, it's important to note a mismatch between the driving sequence provided by the route planner and the actual driving behavior.

Since Route planner is based on the Open-Drive interface and it does not explicitly provide follow lane behavior, instead, Follow Lane behavior is calculated based on the distance to the next driving instruction. This makes it difficult to integrate Follow Lane behavior into the network, which is essential for autonomous driving.  The Follow Lane behavior relies on the Route Planner’s distance to the next driving instruction and can be calculated from there. For example, if a left turn is planned in 100 meters, but the neural network was trained to handle turns starting only 5 meters ahead, it will have difficulty managing the 95 meters it wasn’t trained for. The key variable here is the Turn Distance, which should match the distance used during training. For instance, if training began 5 meters before a turn, the Turn Distance should also be set to 5 meters. When the vehicle is farther than the Turn Distance, the default Follow Lane behavior is used to maintain stability. Once the vehicle is within the Turn Distance, the specific behavior for turning is applied. The following section will discuss the proposed approaches in detail.

\subsection{Basic Behavior Selector (no transition and no interpolation): }
A basic Behavior Selector based on route planner inputs is a straightforward approach where the Behavior Selector chooses behaviors directly based on route planner inputs without considering transitions or interpolation. It’s proven to be unsuccess as it doesn’t take into account the vehicle state mentioned in the above section (failed follow-to turns make the vehicle unstable)

\subsection{Transition-Based Behavior Selector:}
\begin{figure}
\centering
\includegraphics[width=0.30\textwidth]{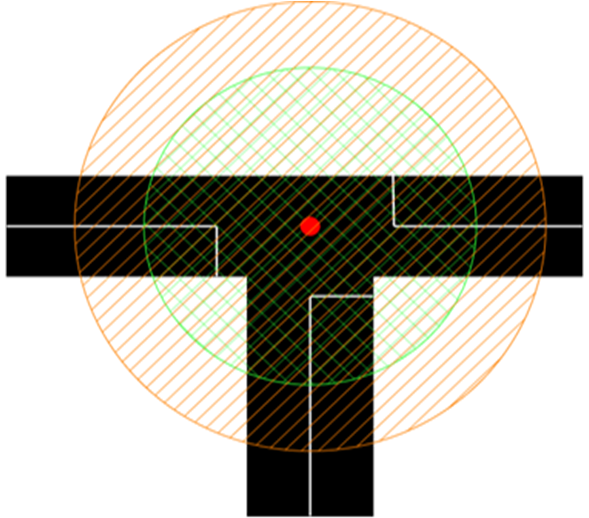}
\caption{Depiction of the turn (green color) and transition (orange color) distance }\label{turn_distance}
\end{figure}

Transition behavior bridges the gap between the vehicle's actual state and the expected state before a turn. It uses a rule-based speed output calculation to ensure safe deceleration. While Follow Lane aims to drive at the maximum safe speed, Transition behavior reduces the vehicle's speed to 1 meter per second (about 3.6 km/h) to meet the requirements for turn behaviors Figure \ref{turn_distance} illustrates Transition Distance added to Turn Distance, showing where the vehicle decelerates before a turn. Transition behavior occurs before turns, not between turns or after turns, as the speed is already low enough after a turn. The Transition Distance starts a few meters before the Turn Distance. 
A static Transition Distance of 3 meters proved suboptimal. Instead, an adaptive Transition Distance based on the vehicle's speed is used.
\[
\text{Transition Distance} = \frac{3 \times \text{speed of the vehicle}}{8}
\]  Transition Distance adjusted according to the vehicle's speed. For example, at 12 \(\text{m/s}^2\), the Transition Distance is 4.5 meters, allowing more time to decelerate. At 4 \(\text{m/s}^2\)
, it's 1.5 meters, reflecting less need for speed adjustment. This adaptive approach ensures smoother transitions and safer driving behavior.
\subsection{Interpolation-Based Behavior Selector}
The other way to tackle the problem with speed is to use a technique called Interpolation.  Interpolation uses both the current and next driving behavior network in parallel, blending their outputs based on the distance to the turn. As the vehicle gets closer to the turn, the influence of the next network increases, helping the vehicle gradually adjust its speed to the expected level for the next maneuver. This approach provides more flexibility than the Transition method, as it avoids unnecessary deceleration by aligning the vehicle’s speed with the next network's output. While Interpolation offers smoother speed transitions, it has drawbacks. It requires more computational power since both networks run simultaneously, and the next network must handle inputs outside its training data. This can lead to unpredictable behavior if the network is not general enough. The computational cost can double, but this might be acceptable depending on the hardware and circumstances. Interpolation coefficient must be effective for any input, that increases as the vehicle approaches the turn, and remain low when the turn is far away. This ensures smooth speed control without unnecessary deceleration when the turn is distant. There are two ways to define interpolation, one is based on interpolation distance and the second is based on turn distance. 
\subsubsection{Interpolation Distance based coefficient}
The first way to define the interpolation coefficient is based on the introduction of the Interpolation Distance. If the vehicle is outside of Interpolation Distance, the coefficient is defined as 0, effectively disabling the interpolation. Inside of the Interpolation Distance the coefficient is defined as the \[
1 - \left(\frac{\text{distance to next behavior}}{\text{interpolation distance}}\right)
\]

\begin{figure}
\centering
\includegraphics[width=0.5\textwidth]{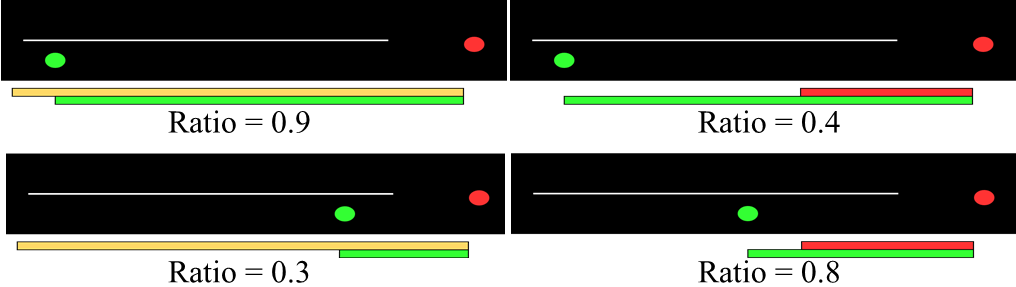}
\caption{Interpolation Distance based coefficient and Turn Distance based coefficient}\label{interpolation}
\end{figure}

This means that as the vehicle gets closer to the next behavior, the distance to that behavior decreases, resulting in a coefficient that approaches 1, thus increasing the influence of interpolation.
Figure \ref{interpolation} on the left side shows a graphical interpretation of the calculations behind Interpolation Distance based coefficient. Red dot represents the coordinate of the next behavior, whereas green dot represents a vehicle. Orange line below the road is the total Interpolation Distance, and the green line is the distance between them. It is easy to see that the ratio between the distances lowers with the vehicle getting closer to the next behavior, which makes the coefficient grow. This is the ease of implementation and understanding however still need to define the Interpolation Distance, which is another variable with a need to optimization.  Due to this reason, I decided to use the following approach for defining the coefficient that is based on the turn distance.

\subsubsection{Interpolation Distance based coefficient}
Another way to define the interpolation coefficient is to use the Turn Distance, a key variable that is already defined in the Behavior Selector. This makes Turn Distance a good baseline for the coefficient function. The Turn Distance based coefficient is calculated as the ratio of the distance to the next driving behavior divided by the Turn Distance. This approach works well when the distance to the driving behavior is greater than the Turn Distance. In regions where the distance is greater than the Turn Distance, interpolation is not applied, making the Turn Distance-Based Coefficient a suitable choice for our needs. Figure \ref{interpolation} (right side) illustrates this concept: green dots represent the vehicle's position, and red dots represent the location of the next driving behavior. The red line indicates the Turn Distance, while the green line shows the distance to the driving behavior as reported by the Route Planner. As the vehicle gets closer to the driving behavior, the distance-to-behavior ratio decreases, increasing the coefficient value. It also simplifies the process by not requiring additional definitions or entities, thus avoiding the complications of alternative approaches. For these reasons, the Turn Distance-Based Coefficient was chosen for the implementation of the Behavior Selector.
\subsection{Behavior Selector based on Transition and interpolation}
As an experiment, the approach combining Transition behavior and Interpolation was also evaluated. The interpolation treats the Transition behavior as an additional behavior to interpolate upon, making the vehicle lower the speed before entering the Transition Distance. After entering Transition Distance, the interpolation proceeds as normal by using output values of Transition Behavior and a behavior corresponding to the next driving behavior. As Interpolation uses the distance to the next behavior, the distance to the Transition behavior should also be provided. The reasonable estimate of the distance to the Transition behavior is distance to the next driving behavior minus the Transition Distance. 
\section{Evaluation and Results:}
This section presents the evaluation of all proposed approaches described in the above section and the comparison among them. The first part of this section provides a detailed overview of the environment and track layout used during evaluation. In the second part of the section, E2E driving network used during the evaluation is explained. After that, a brief overview of route planner and finally results from the evaluation are discussed. 
\subsection{Evaluation Environment:}
\begin{figure}
\centering
\includegraphics[width=0.50\textwidth]{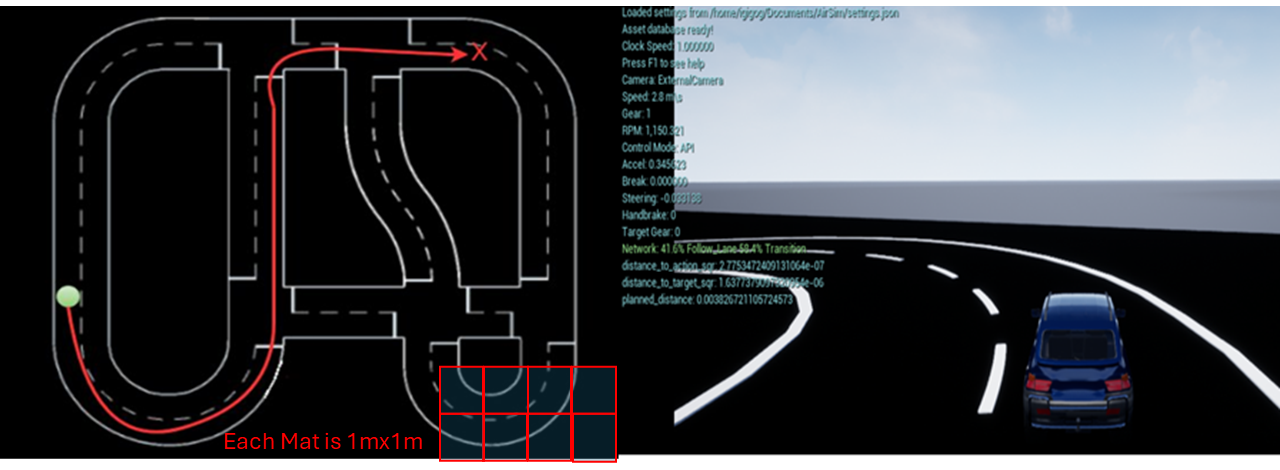}
\caption{Bird view of test track and the vehicle}\label{track}
\end{figure}

The proposed concept was evaluated using a simulation environment created with Unreal Engine \cite{unreal_engine:online} and the AirSim plugin \cite{ros2_docs:online}.  The simulation allows for an efficient and cost-effective assessment, avoiding the delays and expenses of physical hardware. Simulation accelerates the data collection process that would otherwise demand skilled drivers, large training grounds, and extensive supervision. Figure \ref{track} illustrates the simulation environment along with test vehicle and also the bird eye view of test track. 

The test track is designed using a grid of 1m-by-1m mats, each representing different road segments such as straight paths, crossings, or turns as illustrate in figure \ref{track} (left side in the bottom with red block). These mats can be combined to form larger road features, like complex turns or intersections. The training map in the simulation is a composite of these mats, creating a comprehensive road network with various driving scenarios, including turns, straight roads, and crossings with multiple exits. In Figure \ref{track} (left side), the green circle marks the starting position of the vehicle for the test, while the "X" indicates the endpoint. This path was chosen because it is the longest and covers most of the driving behaviors, such as lane following, crossing intersections, and making turns. 

The path for the evaluation is shown in Figure \ref{track} left side. This path was chosen for two reasons. First, it spans over the large part of the track, giving the possibility easier to observe and understand subtle changes in behavior. Secondly, it has multiple points of interest, including large straight lanes, and turns. The reproducibility of the testing round for each proposed approach was ensured by keeping the destination as well as the starting position constant. To keep the starting position constant, the spawn position of the vehicle was placed directly on it, making even small differences impossible. The placement of the vehicle on the starting position was ensured by the computer, as opposed to manual placement made by humans with the possibility of human error. The destination was saved in the form of coordinates and directly loaded into the Route Planner. As there is no change to the saved coordinates, we can be quite certain that both the starting position and destination were constant, leading to reproducible results. There are in total 4 evaluation runs; a) No Transition and Interpolation b) Transition, no Interpolation c) Interpolation, no Transition; d) Transition and Interpolation, with data collection enabled.

\subsection{Driving Behavior networks}
The driving behaviors in this study are primarily implemented using E2E artificial neural networks. These E2E networks process camera data as input and produce speed and steering outputs, with each network tailored to a specific driving behavior. The key driving behaviors implemented include \textbf{\textit{Follow Lane}}, \textbf{\textit{Turn Left}},\textbf{\textit{Turn Right}}, and \textbf{\textit{Cross Crossing}}.

In addition to these, two auxiliary behaviors, \textbf{\textit{Transition}} and \textbf{\textit{Stop}} behaviours, were implemented by overriding the network's speed output with static values. Transition uses a fixed speed of 1.0 m/s², while Stop uses 0.0 m/s². These auxiliary behaviors rely on the steering computations from the Follow Lane network, with the static speed adjustment proving challenging to train due to specific data requirements.

Training data for each driving behavior was collected by manually driving the vehicle in the simulation environment, with recording activated only during the relevant maneuvers to minimize overlap between behaviors. For instance, training data for the Follow Lane behavior excluded turns. Each driving behavior's data was independently recorded and organized into separate datasets. The collection process involved driving to the starting point, recording the specific maneuver, and then saving the data.

\subsection{Discussion of results}

\subsubsection{Basic Behavior Selector }
\begin{figure}
\centering
\includegraphics[width=0.50\textwidth]{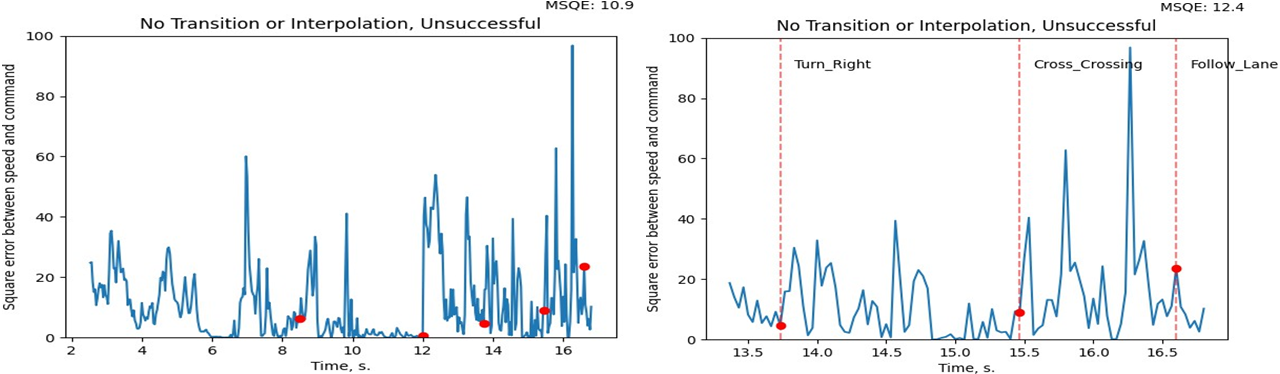}
\caption{Squared error between predicted and actual speed over time for Basic Behavior Selector}\label{basicbs}
\end{figure}

In the Basic Behavior approach, no additional mechanisms were implemented as described in Section 3.1. As a result, failures in the evaluation were anticipated. Figure \ref{basicbs} illustrates the graph of the squared error between the neural network's predicted speed and the vehicle's actual speed over time, with the y-axis representing speed error and the x-axis representing time. The red dots indicate moments when behavior switches occur. It is evident that each behavior switch is followed by a spike in the discrepancy between the predicted and actual vehicle speeds. This outcome is expected because the vehicle's current state (in terms of speed and orientation) does not align with the conditions under which the network was trained.

Looking closely at the behavior switch from the Follow Lane behavior to the Turn Right behavior (Figure \ref{basicbs}, right side), a significant spike in error between the vehicle's current speed and the Turn Right behavior's predicted speed command is clearly observed. This indicates that the Turn Right behavior's driving function interface was not respected. Specifically, the Turn Right behavior failed the maneuver due to two primary reasons: (1) the vehicle's speed was too high to successfully execute the Turn Right maneuver, resulting in failure, and (2) the vehicle was not properly aligned with the lane as required by the Turn Right behavior. Consequently, this evaluation was deemed unsuccessful.

\subsubsection{Transition based behavior selector }
\begin{figure}
\centering
\includegraphics[width=0.50\textwidth]{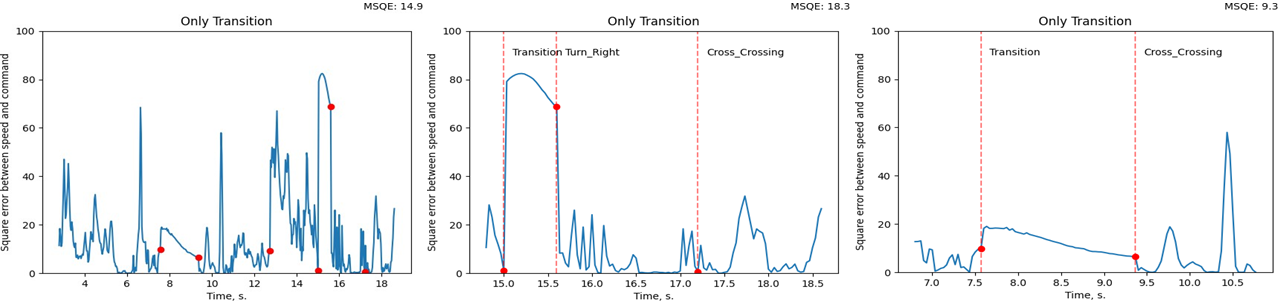}
\caption{Squared error between predicted and actual speed over time for transition based behaviour}\label{transitionbs}
\end{figure}

The integration of Transition behavior enables the Behavior Selector to better adhere to the driving function interface. This implementation is straightforward, requiring no additional training and introducing minimal computational overhead, primarily through simple distance calculations. However, while it offers advantages, the Transition behavior also impacts the overall error between the vehicle's actual speed and the predicted outputs of the driving behavior.
Figure \ref{transitionbs} shows significant error spikes are observed during certain transitions, such as the spike between 15 and 16 seconds. These spikes are due to the synthetic nature of the Transition behavior, which focuses on rapid deceleration rather than accounting for vehicle momentum. A closer look at the largest spike between 15 and 16 seconds, shown in the middle of Figure \ref{transitionbs}, reveals that the error stems from the Transition behavior. The red dotted lines mark the behavior change points, with the subsequent turn right behavior. Despite the spike, this does not indicate a failure, as the evaluation was completed successfully. Importantly, the transition from Transition to Turn Right behavior results in only a modest error, indicating that the vehicle's speed was sufficiently reduced for the turn.

Figure \ref{transitionbs} (right side) also shows a smaller error spike between 7.5 and 9.5 seconds. Again, this spike is due to the Transition behavior, underscoring that while it can increase the error between the vehicle's speed and the neural network's output, it still maintains compliance with the requirement of driving behavior.

\subsubsection{Behavior Selector based on Interpolation behavior}
\begin{figure}
\centering
\includegraphics[width=0.50\textwidth]{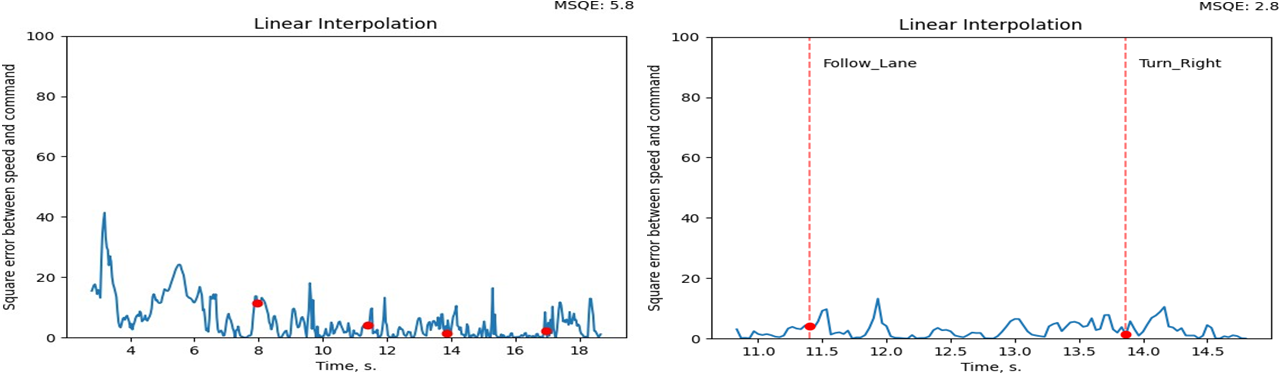}
\caption{Squared error between predicted and actual speed over time for interpolation based approach }\label{interpolationbs}
\end{figure}

The Behavior Selector using Interpolation has demonstrated superior performance in minimizing the squared error between the vehicle's speed and the neural network's predicted output, while also being relatively fast at completing the objective.  Figure \ref{interpolationbs} illustrates the squared error in speed for the Behavior Selector with Interpolation. Notably, behavior switches do not result in error spikes, indicating that the driving function interface is consistently respected. The mean squared error for this configuration is the smallest among all successful runs, achieving an average error of 5.8 m/s², which is a threefold improvement compared to the Transition-based architecture's mean error of 14.9 m/s². 

Figure \ref{interpolationbs} (right side) provides a close-up of the speed error graph around the transition to Turn Right behavior. The absence of a spike during the transition from Follow Lane to Turn Right confirms the effectiveness of the Interpolation technique in maintaining smooth behavior switching.

This configuration offers significant improvements over other approaches, but it has some drawbacks, primarily its computational expense, as it requires running an additional neural network in parallel. Secondly, it has its limitations, particularly concerning steering interpolation. Manual observations revealed that while speed interpolation marginally improved performance, steering interpolation introduced instability into the system. Despite the overall success of the evaluation run, the interpolation of steering values did not enhance stability and, in some cases, detracted from it.

\subsubsection{Behavior Selector based on Transition and Interpolation behavior}
This approach incorporating both Transition behavior and Interpolation resulted in a successful run, though it introduced several notable differences compared to previous approaches. Integrating Transition behavior with Interpolation contributed to increased stability in steering, as the steering commands during Transition were derived from the Follow Lane behavior. This ensured consistency in steering even during interpolation. Additionally, the vehicle's speed was slightly reduced, which is expected given that Transition behavior is designed to slow down the vehicle. Incorporating a portion of this behavior into the Follow Lane command naturally decreased the overall speed.

\begin{figure}
\centering
\includegraphics[width=0.50\textwidth]{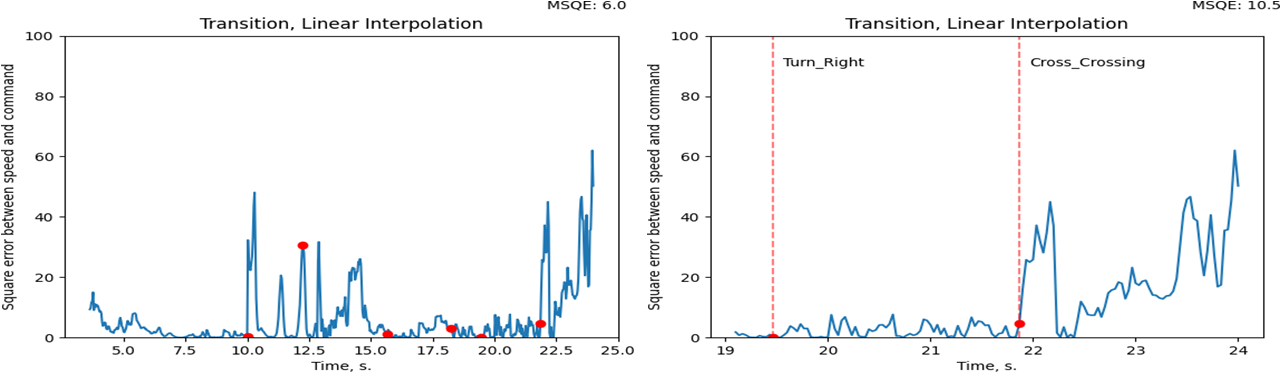}
\caption{Graph of squared error for in cooperation of transition and interpolation approach}\label{hybridbs}
\end{figure}

Figure \ref{hybridbs} presents the speed error, it is evident that the error remained low following behavior transitions, validating the effectiveness of the Interpolation approach. However, a trade-off was observed: the time taken by the system to reach the destination was slightly longer. Specifically, the vehicle took approximately 25 seconds to complete the task, representing a 25\% increase in time compared to both the Interpolation-only and Transition-only architectures.

Figure \ref{hybridbs} (right side) provides a close-up of the error graph at the transition to Turn Right. While the transition to Turn Right did not result in a significant error spike, the subsequent switch to Cross Crossing introduced instability, leading to a notable discrepancy between the actual vehicle speed and the predicted speed command. These results suggest that combining Transition and Interpolation behaviors does not offer significant improvements over using either technique independently or, in some cases, may even exacerbate instability.

\section{Conclusion and outlook }
This paper presents a behavior-based approach to E2E autonomous driving, emphasizing the critical role of the Behavior Selector in managing multiple neural networks tailored to specific driving behaviors. Four distinct strategies were proposed and evaluated to ensure smooth and safe switching between these driving networks: the Basic Behavior Selector, Transition-based Selector, Interpolation-based Selector, and a hybrid model that combines both transition and interpolation.

The evaluation results underscored the significance of transition and interpolation mechanisms in achieving stable and safe vehicle behavior. Runs using a Behavior Selector without these mechanisms resulted in failure, while those that adhered to the transition and interpolation interfaces successfully completed the driving tasks, highlighting the importance of seamless behavior transitions. Among the two approaches—transition and interpolation—both demonstrated effectiveness; however, interpolation provided greater stability while delivering comparable results to the Transition-based architecture, albeit with a significant increase in computational costs.

While the simulation environment offered valuable insights into these approaches, future research will focus on refining and validating these strategies in real-world conditions. This will involve optimizing the Behavior Selector for diverse driving environments and exploring the integration of advanced machine learning techniques to enhance the system's adaptability and robustness, ultimately contributing to safer and more reliable autonomous driving solutions.
 \bibliographystyle{ACM-Reference-Format}
\bibliography{literature}
\section{Acknowledgments}

I would like to acknowledge the contributions of Igor Anpilogov, whose bachelor's thesis provided the foundation for this research paper. I also extend my gratitude to the Institute for Software and Systems Engineering at TU Clausthal for providing essential resources and support during the research process.
\end{document}